\documentclass[letterpaper, 10 pt, conference]{ieeeconf}  

\IEEEoverridecommandlockouts 

\usepackage{acronym}
\usepackage{amsmath}
\usepackage{amssymb} 
\usepackage{caption}
\usepackage{graphicx}  
\usepackage{subcaption}
\usepackage{hyperref}
\usepackage{placeins}
\usepackage{dblfloatfix}
\usepackage{siunitx}

\acrodef{lp}[LP]{Linear Program}
\acrodef{llm}[LLM]{Large Language Model}
\acrodef{uav}[UAV]{Unmanned Aerial Vehicle}	
\acrodef{uavs}[UAVs]{Unmanned Aerial Vehicles}
\acrodef{tacos}[TACOS]{Task Agnostic Coordinator of a multi-drone System}
\acrodef{cot}[COT]{Chain of Thought}
\acrodef{hsi}[HSI]{Human-Swarm Interaction}
\acrodef{icl}[ICL]{In-Context Learning}
\acrodef{mnl}[MNL]{Monolithic}

\title{\LARGE \bf
TACOS: Task Agnostic COordinator of a multi-drone System 
}

\author{Alessandro Nazzari$^{*1}$ and Roberto Rubinacci$^{*1}$ and Marco Lovera$^{1}$ 
\thanks{*Equal contributions}
\thanks{$^{1}$The authors are with Dipartimento di Scienze e Tecnologie Aerospaziali,
		Politecnico di Milano, Via La Masa 34, 20156 Milano, Italy
		{\tt \{alessandro.nazzari, roberto.rubinacci, marco.lovera\}@polimi.it}}%
}

\begin{document}

\maketitle
\thispagestyle{empty}
\pagestyle{empty}

\begin{abstract}
When a single pilot is responsible for managing a multi-drone system, the task may demand varying levels of autonomy,  from direct control of individual UAVs, to group-level coordination, to fully autonomous swarm behaviors for accomplishing high-level tasks.
Enabling such flexible interaction requires a framework that supports multiple modes of shared autonomy. As language models continue to improve in reasoning and planning, they provide a natural foundation for such systems, reducing pilot workload by enabling high-level task delegation through intuitive, language-based interfaces.
In this paper we present TACOS (Task-Agnostic COordinator of a multi-drone System), a unified framework that enables high-level natural language control of multi-UAV systems through \acp{llm}. TACOS integrates three key capabilities into a single architecture: a one-to-many natural language interface for intuitive user interaction, an intelligent coordinator for translating user intent into structured task plans, and an autonomous agent that executes plans interacting with the real world. TACOS allows a \ac{llm} to interact with a library of executable APIs, bridging semantic reasoning with real-time multi-robot coordination. We demonstrate the system on a real-world multi-drone system, and conduct an ablation study to assess the contribution of each module.

\end{abstract}


\section*{IEEE preprint statement}

This work has been submitted to the IEEE for possible publication. Copyright may be transferred without notice, after which this version may no longer be accessible

\section{INTRODUCTION} \label{sec:introduction}

Coordinating multiple \acp{uav} has become a core robotics challenge, with applications ranging from surveillance and mapping to disaster response and delivery.
Assigning a dedicated pilot to each \ac{uav} does not scale with swarm size: it is expensive, inefficient, difficult to coordinate, and prone to human error. As a result, multi-UAV coordination systems must fulfill two critical requirements: provide a high-level intuitive interface for the user, and manage the swarm’s behavior autonomously, while including support for user situational awareness.

Previous approaches have explored a range of interaction modalities. Some systems use fusion modules that combine voice and gesture recognition to interpret commands \cite{mimmo2016}, while others rely on tablet-based interfaces for manual control \cite{wattearachchi2025designing}. However, recent advances in \acp{llm} have opened new directions in autonomous systems, going beyond basic natural language interfaces for human-robot interaction. 

Recent works have explored the use of \acp{llm} to directly generate robot commands from natural language input \cite{kumar2025}. Other approaches have gone further, using LLMs to produce executable code in real time for planning and control tasks \cite{codeaspolicies, zhu2025onlineautomaticcodegeneration}. A particularly compelling direction is the ReAct framework \cite{yao2023react}, which  allows language models to interact with the environment enabling reasoning and action in a closed feedback loop. Building on these ideas, several studies have investigated \ac{llm}-driven control in both single-robot \cite{fan_audere} and multi-agent \cite{fan_scalable_multirobot, roco2023, zhang2024, nayak2024long} scenarios, demonstrating promising results. These developments naturally connect with long-standing challenges in multi-UAV systems, where researchers have developed efficient algorithms for distributed task allocation \cite{ChoiHow2009consensus, Akella2010Consensus} , trajectory generation \cite{atomica} , and formation control \cite{zhou2022swarm, Anderson2016formation}. Many of these algorithms can be exposed as callable APIs, making them well suited for integration with \ac{llm}-based agents under the ReAct framework. 

In this work, we bridge this gap by applying the ReAct paradigm to swarm coordination. We use a language model to interpret high-level user instructions and interact with a library of low-level swarm actions via structured API calls. Specifically we present TACOS, an \ac{llm}-powered coordinator for multi-drone system. Beyond enabling an intuitive one-to-many user interface, a relevant contribution in itself, our framework allows the swarm to benefit from the semantic and commonsense reasoning capabilities of large language models. This paves the way to more flexible and resilient swarm mission execution in unpredictable settings. 

TACOS is designed to support the following core capabilities:
\begin{itemize}
    \item \textbf{One-to-many natural language interface}. Users can issue direct, \ac{uav}-specific instructions such as “\textit{Alfa, take off}”, “\textit{Move Alfa toward the north}” or “\textit{Swap Alfa and Bravo’s positions}”;
    \item \textbf{Intelligent coordinator}. TACOS supports high-level swarm command like “\textit{Split the swarm into two groups, send one group north, and have the other surround the target}”;
    \item \textbf{Task manager}. The system generates structured execution plans to fulfill complex tasks and continuously monitors the swarm’s state to ensure successful execution despite possible drone failures.
\end{itemize}

To the best of our knowledge, this is the first demonstration of a language model interfaced with a real-world multi-drone system for one-to-many interaction and closed-loop task execution.

The remainder of the paper is organized as follows: Section~\ref{sec:problem_setup} introduces the problem setup. Section~\ref{sec:tacos} presents the proposed framework, TACOS. 
Section~\ref{sec:ablation} presents the ablation study performed on the components of the framework.
Section~\ref{sec:Target_search_case_study} and Section~\ref{sec:real_world_exp} reports both simulation results and real-world experiments using quadrotor platforms.

\section{Problem setup} 
\label{sec:problem_setup}
We address the problem of centralized, high-level task planning for a multi-UAV system, via an intelligent coordinator that interprets high-level user commands expressed in natural language. The coordinator must translate these commands into executable plans for the \ac{uav} swarm.

The swarm operates in a bounded 3D environment \(\mathcal{W} \subseteq \mathbb{R}^3\). The environment contains a set of \(n\) ellipsoidal obstacles \(\mathcal{O} = \{ \mathcal{O}_1, \ldots, \mathcal{O}_n \}\), where each obstacle is defined by its center position and a positive definite matrix specifying its orientation and size. The free space is given by \(\mathcal{W}_{\text{free}} = \mathcal{W} \setminus \bigcup_{i=1}^n \mathcal{O}_i\). The environment also includes a set of \(m\) task-relevant entities, such as targets or landmarks denoted by \(\mathcal{T} = \{ \mathcal{T}_1, \ldots, \mathcal{T}_m \}\), each representing a 3D coordinate. The state of the world is thus defined as \(\mathcal{S}_{\mathcal{W}} = (\mathcal{O}, \mathcal{T})\).

We consider \(N\) quadrotor \acp{uav}, where each \ac{uav} has a discrete operational mode \(\mathcal{S}_d \in \{\textit{unavailable}, \textit{idle}, \textit{flying}, \\ \textit{tracking} \} \). The swarm state \(\mathcal{S}_s\) is defined as \(\mathcal{S}_s = \{ (\mathcal{S}_d^{i},p^{i},v^{i} )\}_{i=1}^{N}\), where \(p^{i} \in \mathbb{R}^3\) and \(v^{i} \in \mathbb{R}^3 \) are the current position and velocity vectors of the \(i\)th drone. 
Additionally, each \ac{uav} is equipped with a predefined set of low-level control primitives, formalized as the action set: \(\mathcal{A} = \{ \texttt{arm\_takeoff()},\ \texttt{start\_tracking(}\mathcal{T} _k\texttt{)},\\ \ \texttt{stop\_tracking()}, \texttt{land()}, \texttt{goto(x, y, z)}\}\), which abstract away platform-specific dynamics. 

Each action in \(\mathcal{A}\) is implemented by a dedicated control algorithm that executes the corresponding behavior, thereby reducing the computational and control workload of the \ac{llm}-based framework. For example, the \(\texttt{goto(x, y, z)}\) primitive can invoke a centralized or distributed trajectory generation module to ensure collision-free motion, while 
\( \texttt{start\_tracking(}\mathcal{T} _k\texttt{)} \) relies on sensor-based target tracking algorithms operating at high control rates. The \ac{llm} is therefore not required to run at control frequencies of hundreds of Hertz. Instead, it delegates actions to reliable low-level control algorithms.

This design supports heterogeneous UAV platforms, provided they implement the required action interface. Furthermore, the framework’s capabilities can be easily extended by incorporating additional action primitives, leveraging advances from the wider control and robotics community.

\section{TACOS}
\label{sec:tacos}
\begin{figure}
    \centering
    \includegraphics[scale=0.6]{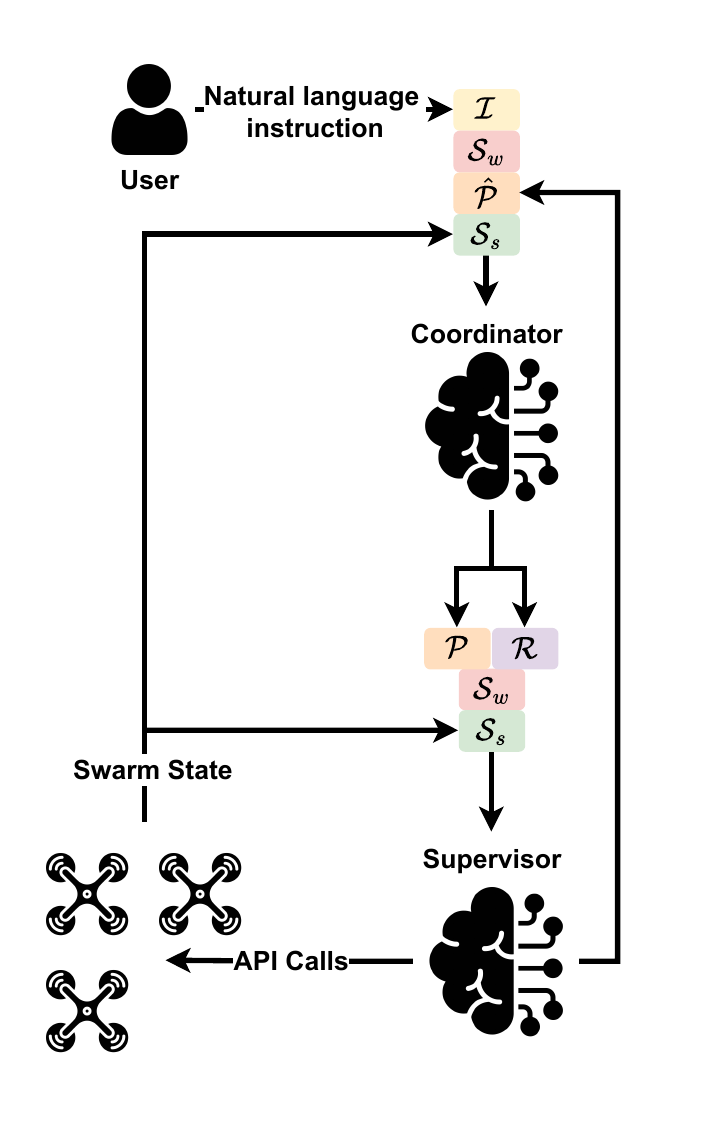}
    \caption{The TACOS framework}
    \label{fig:tacosframework}
\end{figure}

The TACOS framework consists of two main language models arranged in a hierarchical architecture: the Coordinator \ac{llm} which receives high-level natural language commands from the user and synthesizes a task plan, and the Supervisor \ac{llm}, which sequences and executes the plan based on real-time swarm and environment state. 
This hierarchy is inspired by the layered guidance and control architectures commonly used in aerospace systems and by Kahneman two-system theory \cite{kahneman2011thinking}. The Coordinator corresponds to a slow, reasoning layer responsible for planning and intent understanding, while the Supervisor functions as a fast, execution layer. This separation simplifies context management across the \acp{llm} and enables independent customization or fine-tuning of each model.
Each \ac{llm} is initialized with a dedicated configuration prompt that defines its specific objectives, behavioral constraints, and output structure. Figure~\ref{fig:tacosframework} illustrates the modular architecture of TACOS and the interaction flow between user input, Coordinator and Supervisor.

\subsection{Coordinator}
\begin{figure}
    \centering
    \includegraphics[width=1\linewidth]{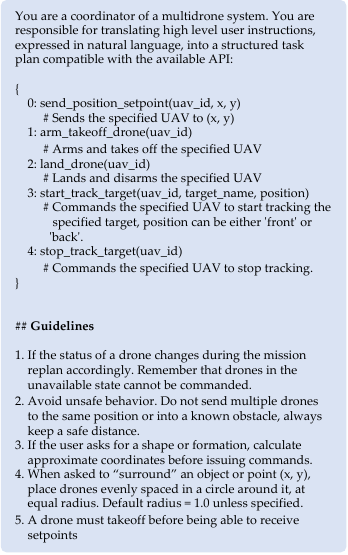}
    \caption{Configuration prompt excerpt}
    \label{fig:available_api}
\end{figure}

The Coordinator \ac{llm} is responsible for translating high-level user instructions, expressed in natural language, into a structured task plan compatible with the swarm’s control interface. 
It receives as input the current system state, comprising the swarm state \(\mathcal{S}_s\), the world state \(\mathcal{S}_{\mathcal{W}}\),  the user’s instruction \(\mathcal{I}\) and the partial task plan \(\hat{\mathcal{P}}\) that encodes unexecuted subtasks from prior plans.
The swarm’s action sets \(\mathcal{A}\) is encoded into the model’s configuration prompt, as shown in Figure~\ref{fig:available_api}. 

The Coordinator’s output is structured into the following two components:

\begin{itemize}
    \item \textit{Reasoning} \(\mathcal{R}\): A natural language explanation of the generated task plan. This explanation supports interpretability. Moreover, prompting the model to explicitly reason improves output quality by leveraging \ac{cot} mechanisms \cite{wei2022chain}.

    \item \textit{Task plan} \(\mathcal{P}\): A list of atomic API calls required to fulfill the user request. All temporal dependencies, synchronization constraints, and inter-agent coordination are deferred to the Supervisor module.
\end{itemize}

An example of the Coordinator's output is shown in Figure~\ref{fig:demo_findthedog}. 

\begin{figure}
    \centering
    \includegraphics[width=1\linewidth]{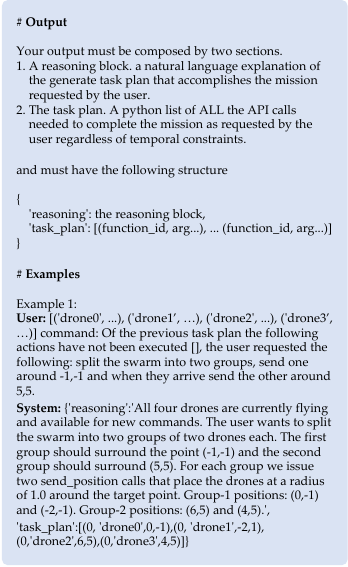}
    \caption{Chain of Thought and In-Context Learning}
    \label{fig:cot_fsp}
\end{figure}

To improve the Coordinator’s planning capabilities, we adopt two prompt engineering strategies: \ac{icl} and \ac{cot}. \ac{icl} allows the model to infer the task structure from a small number of annotated demonstrations provided in the prompt \cite{dong2022survey}. In our case, we provide few-shot examples consisting of user requests and their corresponding ideal reasoning and task plans, illustrating the correct behavior expected from the model. Additionally, we employ \ac{cot} prompting, which instructs the model to explicitly reason through its decisions before generating a final task plan. As illustrated in Figure~\ref{fig:cot_fsp}, the Coordinator is prompted to emit a structured reasoning block followed by the task plan itself. This improves both the coherence of the plan and the interpretability of the decision process.

To enable TACOS to react to dynamic events, the Coordinator is automatically triggered whenever an event requiring a new task plan occurs, for example, when some drones become unavailable or new ones join the swarm. Because only a subset of such events can be automatically detected, TACOS also allows direct user interaction: at any time, the user may query the Coordinator forcing a new task plan to be generated. In this case, the Supervisor informs the Coordinator of the status of the previous plan via \(\hat{\mathcal{P}}\), which records all pending subtasks.

Beyond plan synthesis, the Coordinator also enables high-level swarm control via natural language. It acts as a centralized interface for one-to-many interactions, allowing users to issue commands such as \textit{“Split the swarm and surround the target”}, which are then translated into structured, machine-executable API calls.

\subsection{Supervisor}
The Supervisor \ac{llm} is responsible for transforming the Coordinator’s high-level task plan into a temporal sequence of executable actions. It takes as input the reasoning and task plan produced by the Coordinator, the current swarm state \(\mathcal{S}_s\), the world state \(\mathcal{S}_{\mathcal{W}}\). The set of available low-level actions \(\mathcal{A}\) is encoded directly into the model’s configuration prompt. 

Unlike the Coordinator, which reasons abstractly about intent, the Supervisor operates in a closed-loop execution. Ats each cycle (on the timescale of tens of seconds) it receives updated telemetry from the swarm and the environment and re-evaluates which actions should be issued next. This feedback loop allows the Supervisor to perform temporal sequencing of API calls, respecting dependencies implied by the Coordinator’s plan, and to assign actions to each \ac{uav}.

The Supervisor produces a structured output specifying, for each drone, the current task under execution and the action to be issued. An example of this output is shown in Figure~\ref{fig:sim_6_prompt1}, which illustrates how the Supervisor monitors execution progress and manages action dispatch across the swarm.

To support the coordination between the Supervisor and the Coordinator, TACOS maintains a list of remaining subtasks, denoted by \(\hat{\mathcal{P}}\). This structure captures the portion of the task plan that has not yet been executed and is provided to the Coordinator during replanning events, either triggered automatically by TACOS\footnote{Currently, TACOS triggers a Coordinator replan whenever a drone fails or new drones become available.} or requested manually by the user.

\subsection{History management}
\label{subsec:history}
During a mission, to help the Coordinator understand context and track how the environment changes, the full history of interactions with the user is stored. This includes all past user commands, the corresponding reasoning and task plans, and any updates to the swarm or environment. Keeping this history allows the Coordinator to handle references to earlier commands, \textit{e.g.}, "go back to the last location", and generate plans that are consistent with past instructions. In contrast, the Supervisor \ac{llm} operates with bounded memory. For each task plan, it keeps a temporary record of the commands it has issued and any updates to the swarm or environment. This memory helps it to manage action sequencing, avoid repeating actions, and track progress. Once the current task plan is completed, this memory is cleared, and a new execution cycle begins. This scoped memory model aligns with the Supervisor's role as a reactive executor operating over finite-horizon plans. 

\section{Ablation study} \label{sec:ablation}

\begin{figure}[htbp]
    \centering
    \begin{subfigure}[b]{0.5\textwidth}
        \includegraphics[width=\textwidth]{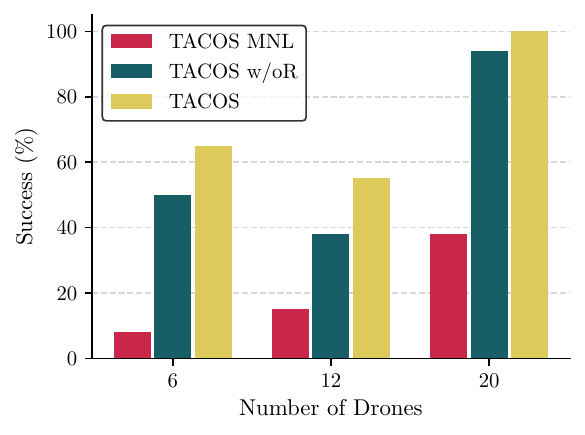}
    \end{subfigure}
    \begin{subfigure}[b]{0.5\textwidth}
        \includegraphics[width=\textwidth]{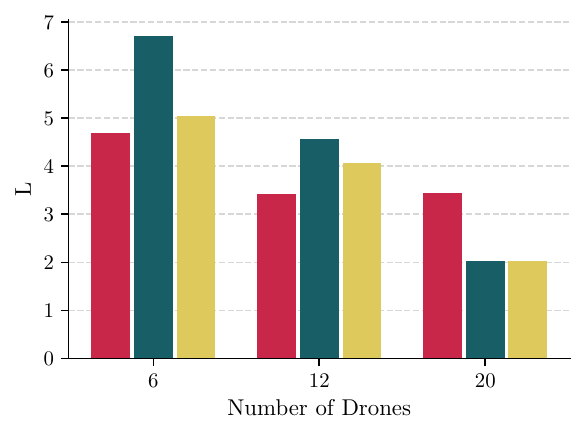}
    \end{subfigure}
      \caption{TACOS performance evaluation}
    \label{fig:Tacos_evaluation}
\end{figure}  
To assess the contribution of each module within the TACOS framework, we perform an ablation study. For each pilot request, we measure the following metrics:
\begin{itemize}
    \item \textbf{Success rate}: the percentage of trials in which the task was completed as intended;
    \item \textbf{Average number of steps} (\(L\)): the average number of Supervisor cycles per task.
\end{itemize}

We evaluate system performance under the following conditions:
\begin{itemize}
    \item \textbf{TACOS Monolithic} (MNL): to evaluate the effect of collapsing high-level reasoning and execution into a single module. In this case, to keep a compact interaction history, we only track executed actions. 
    \item \textbf{TACOS without reasoning} (w/oR): to measure the impact of providing the Coordinator’s reasoning to the Supervisor, and whether it improves the Supervisor’s ability to correctly sequence and complete the task.
    \item  \textbf{TACOS}: The full framework, as shown in Figure \ref{fig:tacosframework}.
\end{itemize}

\subsection{Simulated environment}
\label{subsec:simulated_env}

\begin{figure*}[]      
\centering
\includegraphics[width=\linewidth]{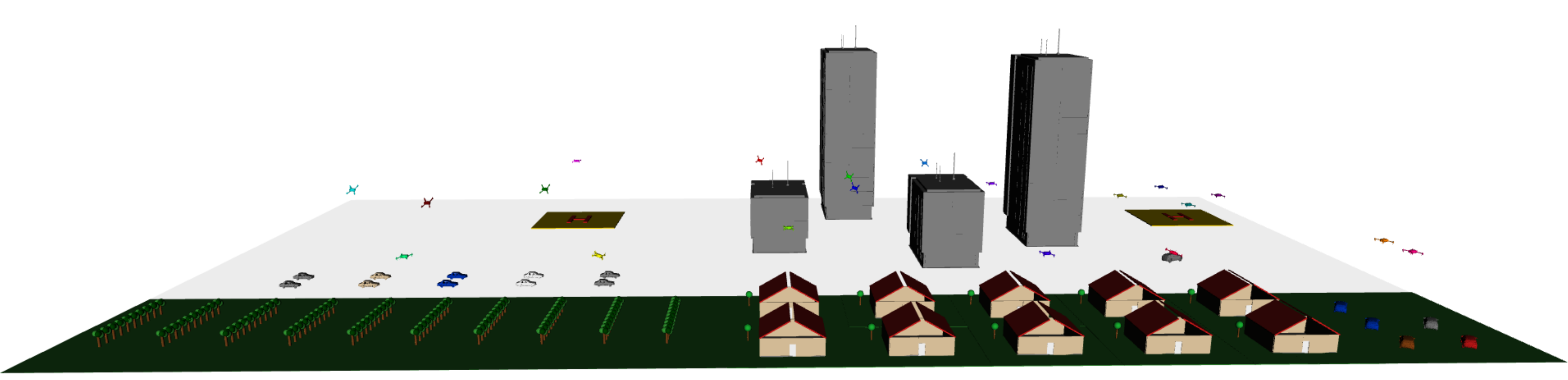}
\caption{Simulation environment}
\label{fig:simulation_environment}
\end{figure*}

The simulation environment, shown in Figure~\ref{fig:simulation_environment}, contains 51 known landmarks, denoted as \(\mathcal{T} = \{ \mathcal{T}_i \}_{i=1}^{51}\), distributed across five distinct areas: two parking lots containing 10 and 5 parked cars, respectively, a residential neighborhood with 10 villas, 
a park area with 20 trees; a business district with 4 skyscrapers; and 2 drone takeoff and landing pads. The four skyscrapers are also modeled as static obstacles to be avoided, represented as \(\mathcal{O} = \{ \mathcal{O}_i \}_{i=1}^{4}\). 

\subsection{Mission}
The mission is structured into two successive tasks designed to test basic capabilities of the framework. 
\begin{itemize}
    \item Task 0: all UAVs are instructed to take off.
    \item Task 1: the pilot requests to find a suspect hiding in a car. Before the mission is completed, a drone failure is simulated to evaluate TACOS response to dynamic events.  
\end{itemize}

For each ablation configuration, we evaluated performance with increasing swarm sizes: 6, 12, and 20 drones. For each setting, we performed 50 simulation runs using the open-source \texttt{gpt-oss} model for both the Coordinator and the Supervisor. The results are summarized in Figure~\ref{fig:Tacos_evaluation}, reporting both the success rate and the average number of steps L, computed over successful trials only. The results for Task 0, \textit{i.e.}, the takeoff task, are not included in Figure~\ref{fig:Tacos_evaluation}, as all configurations successfully completed the task in a single iteration with a 100\% success rate.

The ablation study highlights the following facts:

\begin{enumerate}
    \item Receiving both reasoning and task plan allows the supervisor to infer the correct temporal action sequence. TACOS w/oR has a lower success rate than full TACOS framework considering all three swarm sizes. 
    \item Explicitly dividing task planning and execution helps managing missions requiring multiple execution cycles. This is evident from the superior success rate of TACOS with respect to TACOS MNL shown in Figure \ref{fig:Tacos_evaluation}.
    \item TACOS demonstrates better parallelization capabilities compared to TACOS w/oR. This effect is particularly evident in the number of Supervisor cycles observed with a swarm of six UAVs: TACOS w/oR requires, on average, nearly two additional cycles compared to both the full framework and the monolithic configuration.
\end{enumerate}

In general, we observe that TACOS MNL often issues repeated actions to different drones across iterations, with this behavior worsening as the number of iterations increases. In contrast, TACOS w/oR fails to parallelize tasks, favoring sequential execution. TACOS, on the other hand, does not exhibit these types of failures. It should be noted that, due to the simulated drone failure, the framework cannot complete the task in a single step, even when the number of available drones exceeds the number of targets.

\begin{figure*}[]
\centering
\includegraphics[width=1\linewidth]{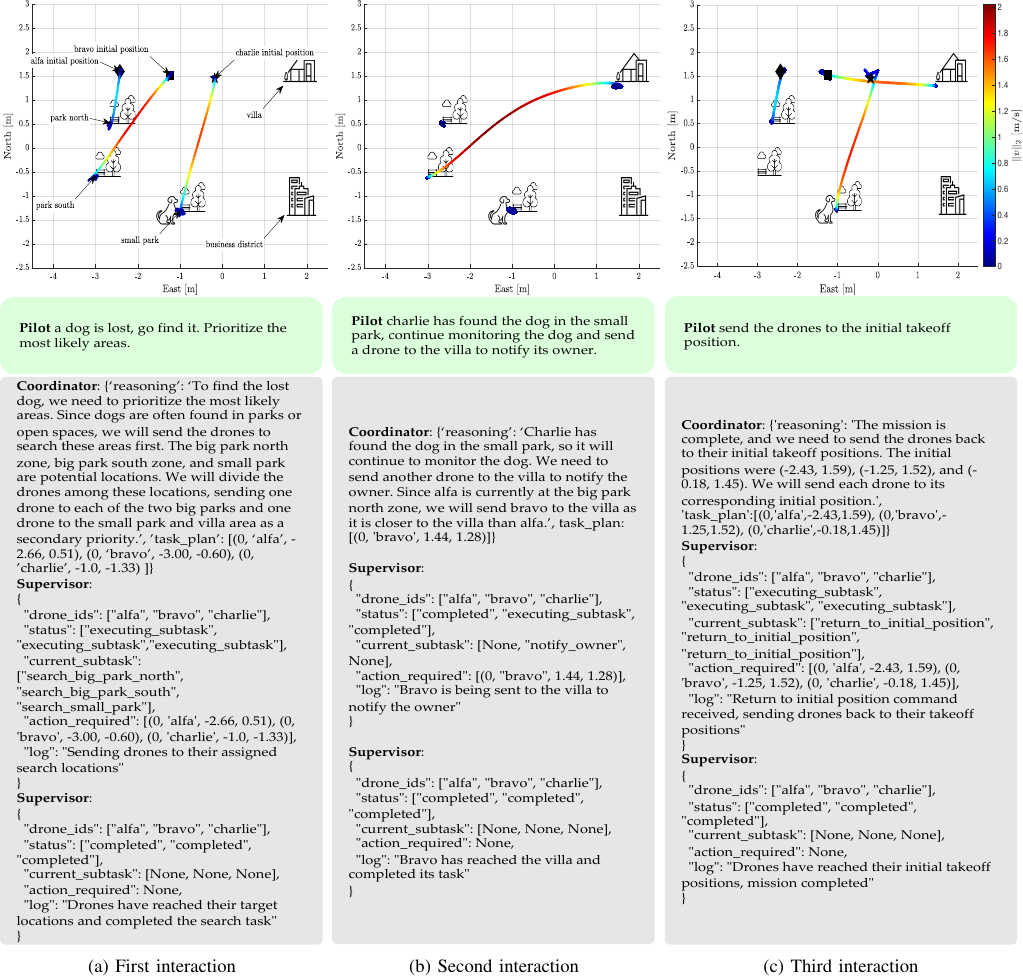}
\caption{TACOS executing the 'Find the Dog' mission by directing drones to the most likely search area based on semantic reasoning.  The figure shows the simplified demonstration environment and the trajectories flown by the drones. Trajectories are color-coded by velocity magnitude, with warmer colors (toward red) indicating higher speeds.}
\label{fig:demo_findthedog}
\end{figure*}

\section{Target search case study}
\label{sec:Target_search_case_study}

This section presents a target search experiment designed to evaluate the performance of TACOS in coordinating a multi-drone system. Specifically, we demonstrate TACOS’s ability to assist a human operator in locating a target within a known environment and subsequently tracking it using multiple drones. We further evaluate the system’s robustness to dynamic events by simulating multiple UAV failures during the mission. 

For real-time trajectory generation, TACOS integrates \textit{ATOMICA}~\cite{atomica}, a real-time, distributed, collision-free multi-agent motion planner. ATOMICA serves as the trajectory generation backend for all \texttt{goto()} and \texttt{start\_tracking()} actions issued by the Supervisor. Experiments have been conducted on a computer running Ubuntu22 LTS using a NVIDIA RTX 6000 GPU. Communication between TACOS and the drones is performed using ROS noetic. The average inference time for the Coordinator is \(9.31\)\si{s} and for the Supervisor is \(5.95\)\si{s}.

The simulated environment is the one described in \ref{subsec:simulated_env}. In addition, the simulation includes a moving target vehicle with known position and velocity, representing a moving agent that attempts to evade the swarm during the search task. For all the simulated experiments, we assume perfect low-level trajectory tracking for the quadrotors. The environment is designed to demonstrate how TACOS assists a human operator through an intuitive natural-language interface, enabling the seamless coordination of large multi-drone systems. The integration of on-board perception and autonomous target detection is left for future work.

\begin{figure}[t]
    \centering
    \includegraphics[width=1\linewidth]{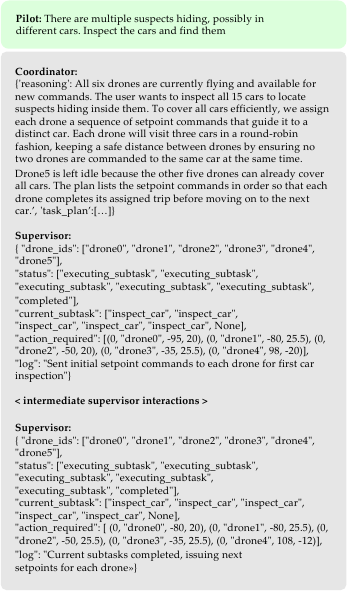}
    \caption{TACOS executing the 'Moving target demo'. Search phase.}
    \label{fig:sim_6_prompt1}
\end{figure}

\begin{figure}[t]
    \centering
    \includegraphics[width=1\linewidth]{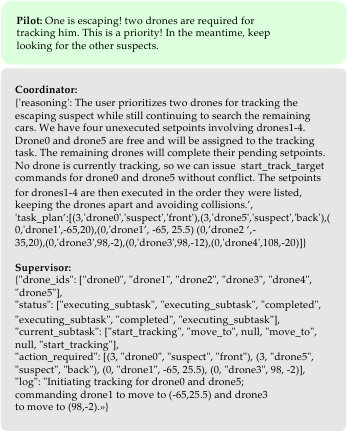}
    \caption{TACOS executing the 'Moving target demo'. Tracking phase.}
    \label{fig:sim_6_prompt2}
\end{figure}

\begin{figure}[t]
    \centering
    \includegraphics[width=1\linewidth]{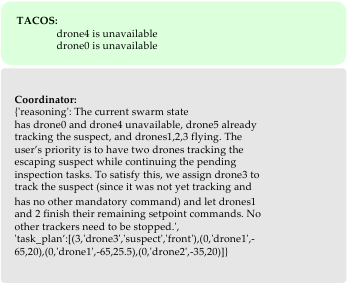}
    \caption{TACOS executing the 'Moving target demo'. Dynamic event.}
    \label{fig:sim_6_prompt3}
\end{figure}

\subsection{Simulation results}
We conducted a series of simulations to evaluate both the capabilities and limitations of TACOS. The mission consists of locating a moving target within the environment and tracking it as it attempts to escape. To assess the ability to handle a dynamic environment, drone failures are intentionally introduced during the simulations, forcing TACOS to replan and reallocate tasks. Videos of the simulation runs are provided in the supplementary material.

We conducted simulations considering swarm of 6, 12 and 20 drones. 
Due to space limitation, we only report here the interactions obtained with a swarm of 6 drones, which highlight the three core capabilities of the framework (see Section \ref{sec:introduction}). Figure~\ref{fig:sim_6_prompt1} shows the initial interaction in which the operator defines the mission objectives through natural language commands. During the execution of the first task, a target escape maneuver is simulated, and the Coordinator is queried to generate a new plan that initiates target tracking.\footnote{No perception module is used, global information about all entities is assumed to be available to TACOS and the drones.} The resulting task plan, shown in Figure~\ref{fig:sim_6_prompt2}, demonstrates that the Coordinator successfully integrates the new tracking request while preserving the pending subtasks from the previous plan. Subsequently, we simulate the failure of two drones, one of which was assigned to track the target. As illustrated in Figure~\ref{fig:sim_6_prompt3}, TACOS triggers a replanning phase, reassigning available drones to complete the mission objectives. 
Leveraging the full history of requests available to the Coordinator, TACOS assigns available drones to each unexecuted task in the partial task plan \(\hat{\mathcal{P}}\) and automatically allocates a second drone to the previously requested tracking task, even without an explicit new command.

It is interesting to note that, in this instance, the Coordinator chose to leave one drone idle during the inspection phase, dividing the inspection of the 15 cars into three consecutive subtasks executed by five drones each. However, the previously idle drone was later assigned to the tracking task. Indeed, the number of drones assigned by the Coordinator may vary across runs due to the inherent stochasticity of \acp{llm}. This variability in the Coordinator’s behavior can be mitigated by carefully shaping the prompt or by fine-tuning the underlying model. To maintain plug-and-play compatibility with different language models, we did not pursue fine-tuning in this work. Nevertheless, we observed that the framework’s behavior can be effectively guided by providing more specific and constrained natural-language instructions to the Coordinator. A key feature of TACOS is its ability to support continuous interaction with the Coordinator, allowing the operator to refine prompts whenever the observed behavior deviates from expectations. For example, when the user issues a new request that interrupts an ongoing task, they can explicitly specify whether the new instruction should take priority over the current one, enabling flexible yet controllable high-level supervision. The demonstration video for all three swarm sizes is available at the following link \href{https://tinyurl.com/TacosSimulated}{https://tinyurl.com/TacosSimulated}.

\section{Real world experiment} \label{sec:real_world_exp}
Flight tests were conducted in an indoor arena equipped with a 12-camera motion capture system tracking markers mounted on the drones. We used three quadrotors running PX4, referred to as alfa, bravo, and charlie. The demonstration environment was a simplified map consisting of a small park, a large park (divided into north and south zones), a villa, and a business district. The mission scenario involved locating a lost dog placed in the small park. Figure~\ref{fig:demo_findthedog} illustrates the pilot’s command and TACOS’s response during the interaction, along with the demonstration environment and the trajectories flown by the drones while executing the commands generated by the Supervisor.

Figure~\ref{fig:demo_findthedog}a highlights the advantages of using TACOS: when the user instructed the system to find the dog, starting from the most likely areas, TACOS inferred, based on semantic reasoning, that parks were more probable locations than the business district. Furthermore, during the initial interaction, TACOS correctly assigned the search task to the closest available drone. In a subsequent interaction, when the user asked to, keep monitoring the dog and notify the owner, the system assigned the task to a farther drone\footnote{In other instances of the same demonstration, TACOS instead selected the closest drone.}.

We repeated the demonstration 10 times. The behavior during the first interaction was consistent across all runs. However, when asked to monitor the dog and notify the owner, the system occasionally chose to monitor the dog with two drones and sent the third one to the villa. The demonstration video is available at the following  \href{https://tinyurl.com/TacosRealWorld}{https://tinyurl.com/TacosRealWorld}.

\section{Conclusion}
In this work, we explored the use of large language models as interfaces for multi-drone systems, and developed a framework that enables a single pilot to control a swarm across a spectrum of shared autonomy. We demonstrated the system in a simplified, lab-scale search-and-rescue scenario, where the semantic reasoning capabilities of the language model improved the efficiency of the search strategy. Additionally, we conducted an ablation study to evaluate the contribution of each module in the proposed architecture. we believe that incorporating onboard perception and expanding the set of available APIs will further enhance the applicability of TACOS.

\bibliographystyle{ieeetr}
\bibliography{bibliography}

\end{document}